# Estimating the Pose of a Euro Pallet with an RGB Camera based on Synthetic Training Data

Posenschätzung einer Europalette mit einer RGB-Kamera basierend auf synthetischen Trainingsdaten

*Markus Knitt*
*Jakob Schyga*
*Asan Adamanov*
*Johannes Hinckeldeyn*
*Jochen Kreutzfeldt*

Institute for Technical Logistics
Hamburg University of Technology

Estimating the pose of a pallet and other logistics objects is crucial for various use cases, such as automatized material handling or tracking. Innovations in computer vision, computing power, and machine learning open up new opportunities for device-free localization based on cameras and neural networks. Large image datasets with annotated poses are required for training the network. Manual annotation, especially of 6D poses, is an extremely labor-intensive process. Hence, newer approaches often leverage synthetic training data to automatize the process of generating annotated image datasets. In this work, the generation of synthetic training data for 6D pose estimation of pallets is presented. The data is then used to train the *Deep Object Pose Estimation* (DOPE) algorithm [1]. The experimental validation of the algorithm proves that the 6D pose estimation of a standardized Euro pallet with a *Red-Green-Blue* (RGB) camera is feasible. The comparison of the results from three varying datasets under different lighting conditions shows the relevance of an appropriate dataset design to achieve an accurate and robust localization. The quantitative evaluation shows an average position error of less than 20 cm for the preferred dataset. The validated training dataset and a photorealistic model of a Euro pallet are publicly provided [2].

*[Keywords: 6D pose estimation, Euro pallet, synthetic training dataset, RGB camera, DOPE algorithm]*

Posenschätzung einer Palette und anderer Logistikobjekte ist von entscheidender Bedeutung für verschiedene Anwendungsfälle, wie automatisiertes Handling oder Tracking. Innovationen in der Bilderkennung, Rechenleistung und maschinellem Lernen eröffnen kamerabasierten Ansätzen auf Basis neuronaler Netze neue Möglichkeiten für die gerätelose Lokalisierung. Hierfür werden große Trainingsdatensätze mit annotierten Posen benötigt. Die manuelle Annotation, insbesondere von 6D-Posen, ist ein äußerst arbeitsintensiver Prozess, weshalb neuere Ansätze oftmals auf synthetischen Trainingsdaten basieren. In dieser Arbeit wird die Generierung synthetischer Trainingsdaten für die 6D-Posenschätzung von Paletten vorgestellt. Anschließend werden die Daten verwendet, um den *Deep Object Pose Estimation* (DOPE)-Algorithmus [1] zu trainieren. Die experimentelle Validierung des Algorithmus belegt, dass die 6D-Posenschätzung einer Europalette mit einer *Rot-Grün-Blau* (RGB) Kamera möglich ist. Der Vergleich der Ergebnisse von drei variierenden Datensätzen unter verschiedenen Lichtverhältnissen zeigt die Relevanz eines geeigneten Datensatzdesigns, um eine genaue und robuste Lokalisierung zu erreichen. Die quantitative Auswertung zeigt für den bevorzugten Datensatz einen durchschnittlichen Positionsfehler von weniger als 20 cm. Der validierte Trainingsdatensatz und ein fotorealistisches Modell einer Europalette sind öffentlich zur Verfügung gestellt [2].

*[Schlüsselwörter: 6D-Posenschätzung, Europalette, synthetischer Trainingsdatensatz, RGB-Kamera, DOPE-Algorithmus]*

## 1  INTRODUCTION

Reliable detection and localization of objects, i.e., the determination of the object's class, as well as position and orientation in space, is crucial to enable flexible automation in complex warehouse and production environments [3]. Localization of assets is required for various logistics use cases, such as asset tracking or automated material handling [4]. Reliable solutions exist e.g. based on Ultra-Wideband (UWB), Bluetooth Low Energy (BLE), or Radio-Frequency Identification (RFID), but require the object to be equipped with a localization device [5]. Due to the high number of assets in a warehouse, such as pallets, and their movement among warehouses and production facilities, device-based localization is often not practical. Furthermore, radio frequency-based approaches only provide positional information. However, many use cases require estimating an object pose, i.e., the position and orientation with up to six degrees of freedom (6D pose).

*Red-Green-Blue* (RGB) and *Red-Green-Blue-Depth* (RGB-D) cameras allow for device-free pose estimation with up to six degrees of freedom. Molter and Fottner [6] apply a 3D point cloud from an RGB-D camera to determine a pallet's pose by identifying the surfaces of the pallet blocks. While the authors show good localization results in the experiments provided, such traditional computer vision approaches often lack flexibility in terms of object class and robustness to occlusion. In recent years, deep learning approaches have proven increasingly effective in solving computer vision tasks and open up new opportunities for camera-based pose estimation. As a major drawback, numerous annotated images are required for training the algorithm. Manually annotating images is a highly labor-intensive task. This is especially true for 3D bounding boxes needed for 6D pose estimation. Mayershofer et al. [7] published an annotated dataset of real-world images for logistics objects that can be used to train algorithms for 2D localization on the image plane. While highly beneficial for training object detection algorithms, estimating the object's 6D pose is not feasible with this dataset. Another source of training data is needed. With synthetic images, large datasets can be automatically generated without the need for labor-intensive manual annotation. Hence, estimating the 6D pose of pallets or other logistics objects with neural networks based on synthetic data is a promising approach. Nonetheless, this has not yet been investigated in the literature. This work aims to fill this gap by providing the following contributions to camera-based 6D pose estimation.

- A photorealistic model of a Euro pallet and a synthetic RGB image dataset with annotations to train 6D pallet pose estimation algorithms [2]
- A pipeline to generate synthetic image datasets and train a neural network for 6D pose estimation with an RGB camera
- A qualitative and quantitative evaluation of the position accuracy and robustness of the trained neural network based on different datasets under varying lighting conditions

The remaining paper is structured as follows. An overview of related work is provided in Section 2. Section 3 describes the pallet pose estimation pipeline. As a first step, it is explained how a photorealistic model of a pallet can be created. Next, the generation of synthetic datasets is described using the *NVIDIA Deep learning Dataset Synthesizer* (NDDS) [8]. *Deep Object Pose Estimation* (DOPE) [1], a state-of-the-art deep learning-based 6D pose estimation algorithm, is finally trained with the generated dataset. Section 4 presents the experiment setup and procedure, as well as the qualitative and quantitative results, which are discussed in Section 5. Finally, conclusions are made and an outlook on future work is provided.

## 2 RELATED WORK

Camera-based pose estimation is a rapidly growing field of technology with a wide range of possible applications. In the following, solutions for pallet detection and pose estimation in the field of intralogistics are presented. Finally, the identified research gaps that are addressed by this work are pointed out.

Molter and Fottner [6] present the pose estimation of a Euro pallet based on an RGB-D camera. First, the number of voxels is reduced by filtering. By detecting surfaces in the point cloud, the three wooden blocks at the side of a pallet and their geometrical relationship are used to determine its 6D pose, under the condition that the pallet is positioned flat on the floor. Empirical experiments show a high repetition accuracy in the range of centimeters with 15fps (frames per second). However, the absolute position error is not evaluated. As a major drawback, this approach lacks flexibility in terms of the object class and robustness to occlusion and different angles.

Domain-specific image datasets have been published to accelerate camera-based object detection in warehouse and production environments. The *Logistics Objects in COntext* (LOCO) dataset [7] provides real images of logistics objects such as forklifts, pallet trucks, pallets, small load carriers, and racks in real warehouse and production environments. The images are captured with an RGB-D camera. Of the 39101 images, 5593 are manually annotated with 2D bounding boxes. In addition, depth maps are available. While this is a suitable dataset for training 2D object detection algorithms, it does not provide 3D bounding boxes of the objects, making it less useful for 6D pose estimation.

Another open-source dataset, called *Synthetic Object Recognition Dataset for Industries* (SORDI) [9], has recently been announced by the *BMW Group*, in cooperation with *Microsoft*, *NVIDIA*, and *idealworks*. Accordingly, SORDI will be made available on *GitHub* within 2022 and will feature over 80 different object classes from the industrial environment, including many logistics objects such as containers, storage boxes, and pallets. The dataset is announced to contain more than 800 000 synthetic photorealistic images that can be used to train vision-based algorithms. Currently, it is unclear if the dataset will include 3D bounding boxes required for 6D pose estimation.

The analysis of the literature reveals two main research gaps. First, despite its potential, camera-based pose estimation for logistics has not been satisfactorily addressed in the literature. This is particularly the case with RBG camera-based 6D pose estimation and with networks based on synthetic training data. Second, publicly provided training data for 6D pose estimation of Euro pallets is missing. This work aims to fill these research gaps by providing the contributions mentioned above.

## 3 PALLET POSE ESTIMATION

This section presents how 6D pallet pose estimation based on synthetic training data can be achieved with an RGB camera. For this, a pipeline to generate an annotated image dataset and finally train a neural network is presented. The pipeline is elaborated using a Euro pallet, since it is one of the most common and standardized objects in warehouse and production environments. However, the method can be adopted for similar object classes. Figure 1 provides an overview of the three steps required for training the network. First, a photorealistic 3D model of a Euro pallet is created. A synthetic image dataset is then generated by applying domain randomization and physically based rendering. In this work, the dataset generation tool NDDS is used to create different datasets. The datasets are finally used to train DOPE. In addition, it is described how DOPE estimates an object's pose.

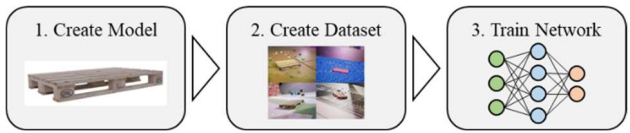

*Figure 1: Pipeline to generate dataset and train network for camera-based 6D pose estimation.*

### 3.1 PHOTOREALISTIC PALLET MODEL

Naturally, the quality of a pose estimation algorithm, based on neural networks that are trained on synthetic training data depends on the underlying model of the object whose pose is to be estimated. Inaccuracies in the appearance, such as shape, size, and material, between the model and the real object can result in poor detection and localization performance. There are different types of pallets that differ in their appearance. The pallet type that is the subject of this work is the EPAL Euro pallet (EPAL 1), which conforms to the requirements described in DIN EN 13698-1 - the norm describing the construction specification for 800mm $\times$ 1200mm flat wooden pallets [10]. Since August 2013, all EPAL pallets must have EPAL markings on both the left and right corner blocks. In addition, for import and export of wood, the International Plant Protection Convention (IPPC) marking is mandatory on the long sides of the pallet on the middle block. When creating the 3D pallet model, the typical markings were considered in addition to size and shape.

A photorealistic model more closely resembles the reflective properties of a real object and therefore results in a better network performance. For creating photorealistic pallet textures, a standard *Physically Based Rendering* (PBR) texture creation workflow was applied (Figure 2). First, high-resolution images were taken of all sides of an EPAL Euro pallet. The images are cropped, leaving only the pallet surfaces. After removing lens distortion effects and adjusting the size, the images were stitched together into a diffuse texture map using image editing software such as *GIMP*. The open-source tool *Materialize* [11] was used to create different texture maps, each of which describes different aspects of the material. Finally, *Blender*'s [12] shader editor was used to combine the different texture maps into realistic materials for wooden planks, wood-chip blocks, EPAL brandings, and IPPC brandings. The textures are assigned to the surfaces of the 3D pallet model. Figure 3 shows a rendered pallet model with the PBR textures applied.

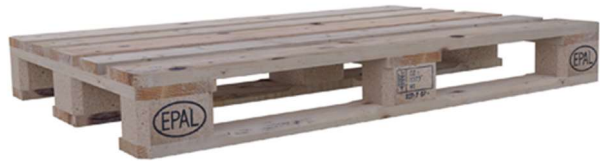

*Figure 3: Rendered pallet model with PBR textures.*

### 3.2 SYNTHETIC IMAGE DATASET

Although there are many aspects of Euro pallets that are standardized, there are still differences in appearance due to variations in wood color and grain, wear-and-tear, dirt, and lighting conditions. Additionally, pallets are placed in different environments with different types of backgrounds. This requires that pallet detection and localization algorithms have a good generalization ability to perform well in practical applications. Without considering these factors, pose estimation algorithms trained on synthetic data tend to perform poorly on real data. Domain randomization is a technique that has been proven successful in improving the generalization ability of deep learning algorithms and bridging the so-called reality gap [13]. With domain randomization, a variety of simulated environments with randomized properties are created. This approach was used for the dataset generation of this work.

Three datasets (NDDS1, NDDS2, NDDS3) were created using NDDS. Figure 4 shows four exemplary images from NDDS3. The created images show scenes containing the previously created photorealistic pallet model. The pallets are automatically annotated with their 6D pose in relation to the camera's coordinate frame. The properties of the three datasets are summarized in Section 3.2. During the dataset generation procedure, the camera orbits around the pallet within a space that is defined by an azimuth range, an altitude range, and a target distance range. Due to the symmetry of the pallet, an azimuth range between -90° and 90° is sufficient. The altitude and distance range represent a scene where the pallet is positioned more or less horizontally on the ground and is viewed obliquely from the side by the camera. This scene represents a typical logistics scenario where a stationary pallet is approached by a logistics vehicle, such as a pallet truck, while the pallet remains fully in the camera's field of view. Domain randomization was implemented by randomizing the lighting, the background textures, and the pallet textures in the first dataset. The datasets were generated and prototypically tested in such a way that the specification of the parameters of a dataset is influenced by the tests of the previous datasets. NDSS2 was driven by the aim of generating an increased number of images with more realistic scenes. For NDDS3 additional effects were added by spawning random objects called distractors and introducing a camera wiggle. The camera wiggle is added to the orbital motion and results in the pallet being captured from different angles.

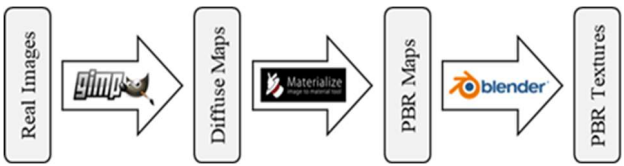

*Figure 2: Workflow for creating a photorealistic model.*

*Table 1. Overview of generated datasets.*

|  | NDDS1 | NDDS2 | NDDS3 |
|---|---|---|---|
| **Nr. of Images** | 50 000 | 100 000 | 50 000 |
| **Azimuth Range** | [-90°, 90°] | [-180°, 180°] | [-90°, 90°] |
| **Altitude Range** | [1°, 35°] | [5°, 25°] | [1°, 25°] |
| **Distance Range** | [3.5m°, 4m°] | [3m°, 3.5m°] | [2.5m°, 5.5m°] |
| **Pallet Textures** | Random colors and patterns | Realistic | Realistic |
| **Lighting** | 3 Random lights and a skylight | 3 Random lights and a skylight | 8 random lights closer to the pallet |
| **Background Textures** | Random colors and patterns | Random colors and patterns, real images | Random colors and patterns, real images |
| **Distractors** | No | No | Yes |
| **Camera Wiggle** | No | No | Yes |

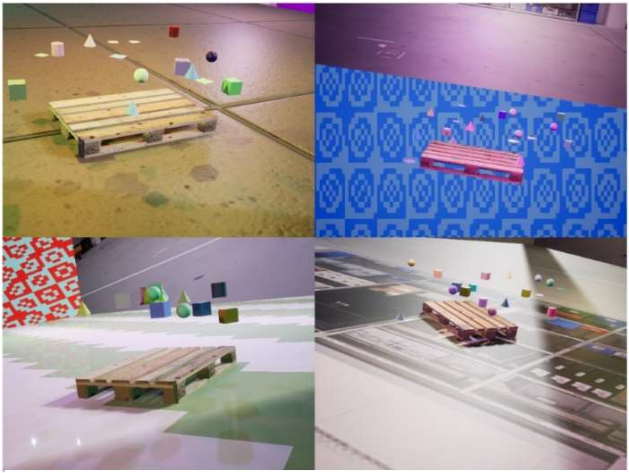

*Figure 4: Example images of the generated data set (NDDS3).*

### 3.3 DOPE - DEEP OBJECT POSE ESTIMATION

As the final step in the pipeline shown in Figure 1, the generated datasets are used to train the neural network. In this work, DOPE was chosen because, unlike many other cutting-edge 6D pose estimation algorithms, DOPE uses solely RGB images to estimate the position and orientation of objects in space.

To train DOPE the datasets were individually loaded by a *Python* script. The datasets are split into training and testing data, where for each dataset 90 percent of the data is used for training and 10 percent for testing. DOPE is trained for 60 epochs, whereby there is one set of weights being generated for each epoch. The weights are saved in *PTH* files - a file type native to the machine learning framework *PyTorch*. The set of weights that produce the smallest loss during testing is chosen for the empirical evaluation. The weights for NDDS3 are provided with the dataset [2].

The network structure of DOPE is shown in Figure 5. The trained network takes an RGB image as an input and outputs a 3D bounding box. It uses VGG19 [14], a pre-trained neural network, to reduce the computational effort. A deep neural network estimates the location of the 2D key points of the objects in the image coordinate system based on an RGB image of size w×h×3, whereby w and h are the width and height in pixels. In this work, a camera is used with w = 640px and h = 480px. The network consists of six stages, where each stage uses the image features as well as the output of the previous stage as input. The network branches and creates two types of outputs. The first being nine belief maps and the second being sixteen vector fields. Eight of the nine belief maps represent each the probability of a vertex of the 3D bounding box. The ninth belief map represents the probability for the centroid of the 3D bounding box. The peaks of the nine belief maps are serving as the input for a standard perspective-n-point (PnP) algorithm [15]. The PnP algorithm then estimates the pose in 3D for each instance of an object. For this, DOPE must be provided with a 3D model of the object of interest. The vector fields indicate the direction from the eight vertices to the respective centroid of the cuboid to allow multiple instances of one object to be detected and localized. The algorithm is explained in more detail in [1].

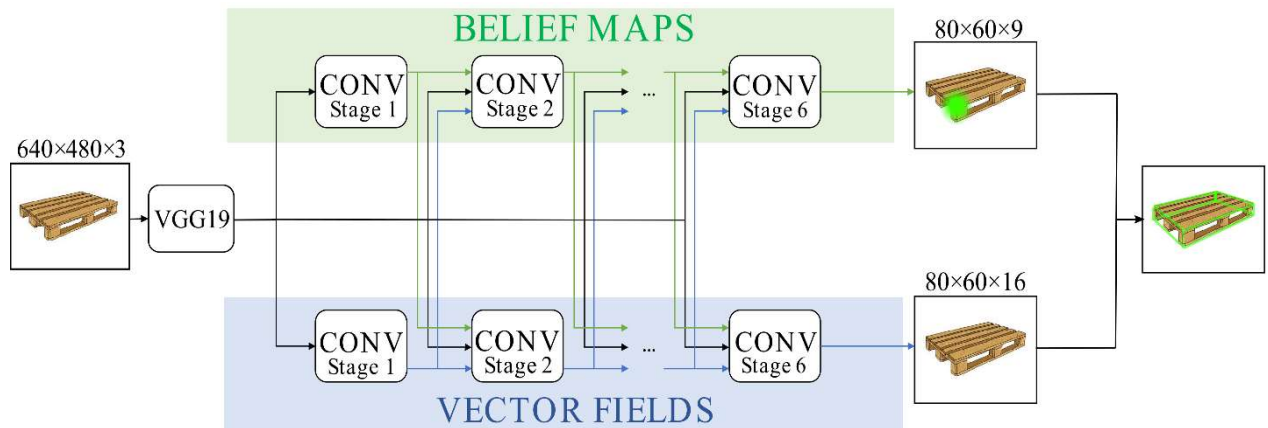

*Figure 5: Network structure of DOPE.*

## 4 EXPERIMENTS AND RESULTS

The trained network with the weights from the NDDS1, NDDS2, and NDDS3 datasets is evaluated based on a series of three dynamic experiments with a low-cost webcam (Logitech C270 HD). The robustness and the position accuracy of the trained algorithm is analyzed based on three experiments under different lighting conditions. The experimental setup and procedure, as well as the qualitative and the quantitative results are presented in the following. For qualitative evaluation, the ground truth positions of the camera are compared to the inverted positions from DOPE. For quantitative evaluation, two metrics are applied to evaluate the position accuracy and robustness for each of the combinations between the three datasets and the three experiments.

### 4.1 EXPERIMENTAL SETUP AND PROCEDURE

The experiments are visualized in Figure 6. The ground truth pose of the webcam and the euro pallet are determined by an optical passive motion capture system (MoCap) similar to the one analyzed by Bostelman et al. [16]. The system from Qualisys reaches position accuracies in the range of millimeters with an update frequency of 100fps. It consists of multiple infrared (IR) cameras mounted on the sides of the test hall. The coordinate system of the MoCap ($w_x$, $w_y$, $w_z$) is referred to as the global coordinate frame. The pallet is positioned in the global origin, by placing a device with IR reflectors on top of the pallet before the experiments (Figure 7). The long side of the pallet is parallel to $w_x$.

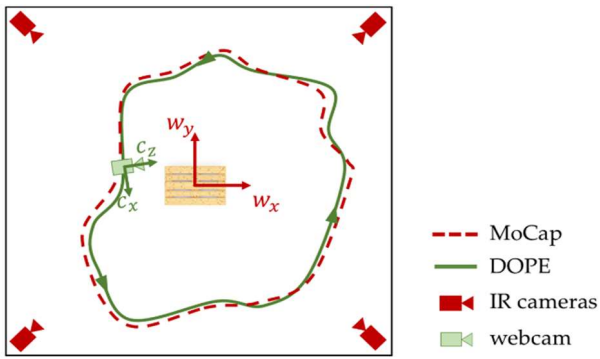

*Figure 6: Visualization of experiments from above.*

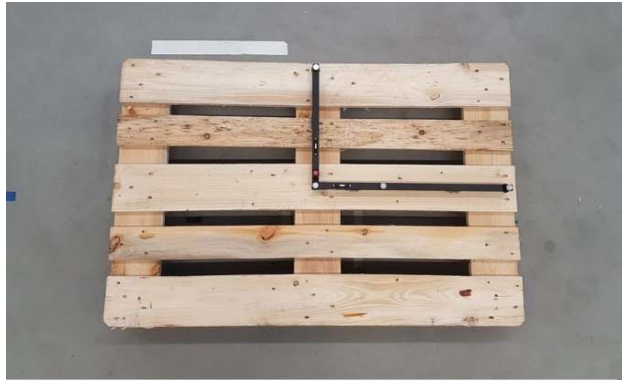

*Figure 7: Positioning of the pallet and definition of global coordinate frame with IR reflectors.*

The webcam that provides the input RGB images for the DOPE algorithm is mounted on a wheeled table along with multiple IR reflectors. The camera's right-handed coordinate system is defined by $c_x$, $c_y$, $c_z$. The camera is tilted towards the ground with a constant pitch of -13.8°. During experimentation, the ground truth pose of the camera is continuously tracked by MoCap. The ground truth trajectory, resulting from the motion of the webcam, is marked by the red dashed line in Figure 6.

The webcam is connected to a workstation containing two graphics cards (NVIDIA GeForce RTX 2080 Ti). From within a Docker container, the ROS (Robot Operating System) nodes are run for the camera and for DOPE. The webcam is moved around the stationary pallet while pointing in the direction of the pallet and maintaining a horizontal distance of approximately 3.5m. The images captured by the camera are stored in a *rosbag*, a ROS-native file format that allows playback of the same sequence of images. The same image sequence can thus be used for the individual evaluation of the network, trained with each of the three datasets.

DOPE publishes the timestamped pose of the pallet with respect to the webcam's coordinate frame in real-time. The time to calculate a position update is distributed either around 0.17s or around 0.21s, resulting in an average update frequency of approximately 0.85fps. Since the pallet is positioned in the global origin, the inverted pose describes the webcam's pose in global coordinates. A pallet is a symmetrical object. For every possible pose there exists another pose from which it cannot be distinguished by DOPE. This needs to be compensated to match DOPE's pose estimates with the ground truth measurements. Thus, 180° is added to the estimated yaw angle of the pallet when changing the direction from which the webcam is facing the pallet. The respective trajectory based on the estimation of DOPE is visualized as the green line in Figure 6.

*Table 2.  Overview of lighting conditions in experiments.*

|  | **Experiment 1** | **Experiment 2** | **Experiment 3** |
|---|---|---|---|
| Artificial Light | Yes | Yes | No |
| Natural Light | Yes | No | Yes |

To analyze the performance of the algorithm trained on the three data sets, the experimental procedure is repeated three times under different lighting conditions, as shown in Table 2. Artificial light means that the ceiling lights are switched on, and natural light means whether the side windows of the test hall are darkened or not. Moving the camera around the pallet has the additional effect that the background is constantly changing. In conclusion, the three experiments, each tested with the weights of the three training datasets, allow the algorithm's performance to be evaluated for different lighting conditions and varying backgrounds.

### 4.2 QUALITATIVE RESULTS

In the following, the qualitative results are presented. Although the output of DOPE is the pose of the pallet with respect to the camera coordinate frame, for qualitative evaluation, the inverse position is considered, i.e., the position of the camera with respect to the global coordinate frame. The representation of the position on the horizontal plane allows for an intuitive interpretation and comparison of the camera's position estimate by DOPE with the ground truth position of MoCap. In addition, position-dependent effects resulting from certain backgrounds or lighting conditions can be analyzed. Figure 8 shows the ground truth and the inverted positions as estimated by DOPE based on the NDDS1, NDDS2, and NDDS3 datasets for the three experiments.

The MoCap points show an approximately circular motion around the origin, i.e. the position of the pallet, at a distance of three to four meters. Matching points of DOPE with MoCap points indicate a successful pose estimation. On the other hand, gaps indicate that DOPE was not able to localize the pallet correctly or to detect the pallet at all, which is referred to as robustness. For all three experiments, most points of NDDS1 match the MoCap measurements, indicating a good accuracy of the pose estimation. However, there are several large gaps, as a result of failed detection of the pallet. NDDS2 shows a higher number of outliers and systematic errors, i.e., longer sequences of points apart from the MoCap points, in all three experiments. The points from NDDS3 indicate the best results in terms of accuracy and robustness. The comparison of the experiments shows a similar ground truth trajectory and similar behavior of the DOPE algorithm in general. Nonetheless, certain effects can be observed. First, NDDS2 shows an increased number of systematic errors for Experiment 2 (only artificial light). Second, for Experiment 3 (only natural light) NDDS1 and NDDS2 indicate a comparably poor robustness. Third, Increased scattering and systematic errors of the DOPE estimates occur around the lower left of the circle in all three experiments.

Figure 9 to Figure 11 aim to provide further insight into the behavior of the algorithm in different challenging scenes. The scenes show challenging lighting conditions (Figure 9), occlusions (Figure 10), or complex backgrounds (Figure 11) for each of the three datasets. In the upper part of the Figures, a green bounding box is shown with the pallet when the pallet is successfully detected. If the pallet is not detected, only a coordinate frame is shown that is corresponding to the last estimated pose. At the bottom of each figure, the nine belief maps are superimposed over the RGB image. As explained in Section 3.3, the belief maps indicate where the algorithm believes each vertex or centroid is and how confident it is of that fact. Ideally, there should only be one clear white dot on each belief map.

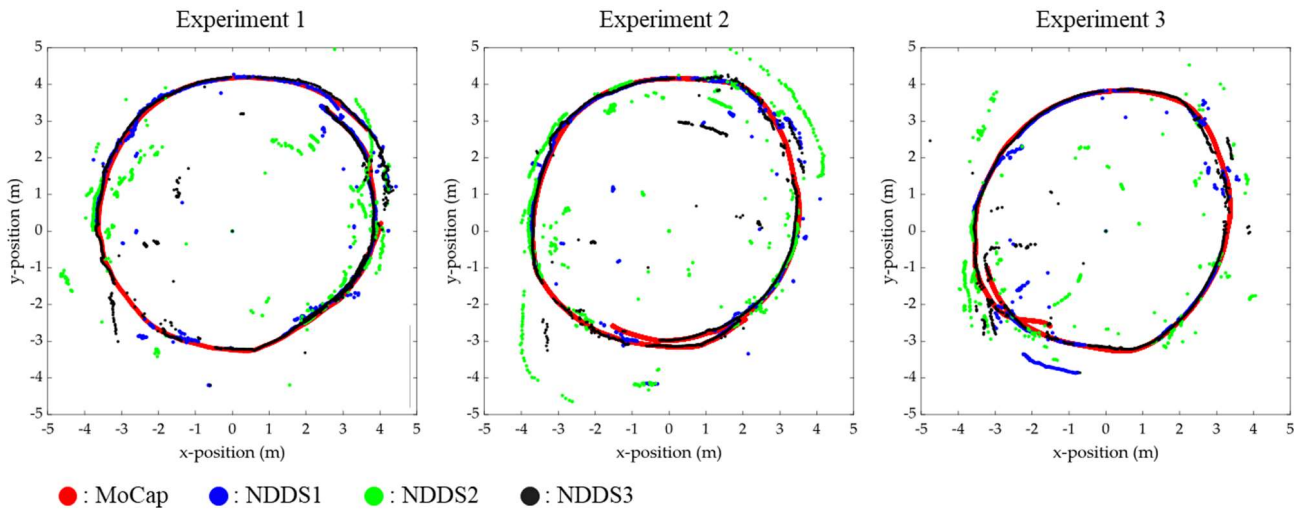

*Figure 8: MoCap and inverted positions as estimated by DOPE with NDDS1, NDDS2, and NDDS3 for Experiments 1, 2 and 3.*

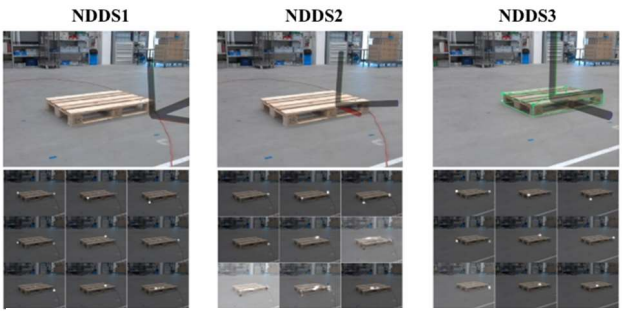

*Figure 9: Robustness to light.*

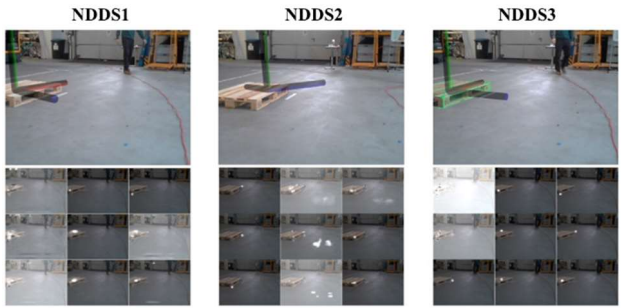

*Figure 10: Robustness to occlusion.*

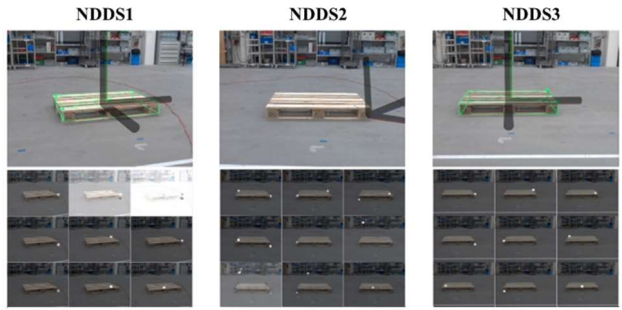

*Figure 11: Robustness to background.*

For the scene in Figure 9, a small window lets in light, resulting in a special lighting situation that affects pose estimation. The camera's position corresponds to the lower left of the circles in Figure 8, where the qualitative results show systematic errors of NDDS1 and NDDS2.Figure Occlusions were not intentionally induced in the experiments. Nonetheless, there are situations, in which the pallet is not fully in the camera's field of view, and thus certain parts are hidden. An example is shown in Figure 10. Finally, Figure 11 shows a scene, in which the shelving rack in the back, creates a complex background. For NDDS2 in particular, DOPE estimates that some vertices of the pallet are inside the shelf, resulting in failed detection. In all three challenging scenes, DOPE trained with NDDS3 is able to detect the pallet successfully.

### 4.3 QUANTITATIVE RESULTS

To allow a quantitative evaluation of the network performance for the different datasets and experiments, the following two metrics are introduced.

(1) The **position accuracy** is determined as the mean 3D euclidian error between the estimated position of the pallet's centroid to its real position by using the *evo* tool [17]. The timestamped pose messages from the MoCap and DOPE are loaded from the rosbag, synchronized by time, aligned, and finally matched to determine the mean euclidian distance of the positions in 3D. Additionally, a discrete-time low-pass filter is applied to the pose estimates of DOPE to reduce the effect of outliers.

(2) The duration of a pose update from DOPE is distributed with two peaks around approximately 0.17s and 0.21s. For a duration of more than 0.3s it is assumed that DOPE does not detect the pallet successfully. The ratio of time in which the pallet is successfully detected to the total measurement time is considered as **robustness**.

Table 3 shows the mean position error of the unfiltered and filtered datapoints and the robustness for all combinations of experiments and datasets. Green numbers indicate the three best and red numbers the three worst results in each column. Filtering reduced the mean positional error in seven out of nine cases. Comparing the mean position error of the datasets with another, NDDS2 shows significantly worse results. For all cases, NDDS1 and NDDS3 show a mean position error of less than 22cm. Comparing the position error of the experiments with different lighting conditions does not show clear differences in the algorithm's performance. However, NDDS2 shows significantly worse robustness of 35.0% in the presence of natural light. For NDDS3 the robustness in all experiments is far superior to the robustness of all the other cases with a minimum of 94.9%.

*Table 3. Quantitative results of the different combinations of datasets and experiments (Exp.).*

| Dataset | Exp. | Position Error (cm) | | Robustness (%) |
|---|---|---|---|---|
| | | Unfiltered | Filtered | |
| **NDDS1** | 1 | 9.7 | 15.2 | 70.7 |
| **NDDS1** | 2 | 15.9 | 8.8 | 62.9 |
| **NDDS1** | 3 | 21.3 | 21.2 | 59.5 |
| **NDDS2** | 1 | 64.4 | 52.8 | 52.0 |
| **NDDS2** | 2 | 66.0 | 59.1 | 73.9 |
| **NDDS2** | 3 | 94.9 | 62.0 | 35.0 |
| **NDDS3** | 1 | 19.2 | 12.9 | 94.9 |
| **NDDS3** | 2 | 16.4 | 17.8 | 96.0 |
| **NDDS3** | 3 | 16.1 | 14.0 | 97.5 |

## 5 DISCUSSION

The presented results show that 6D pallet pose estimation based on synthetic training data with a low-cost RBG camera is feasible. The quantitative results confirm the indications shown by the qualitative results. The DOPE algorithm trained with the NDDS1 dataset shows good performance in terms of position accuracy but only moderate robustness. NDDS2 was designed to increase robustness by generating an increased number of images with more realistic scenes. Compared to NDDS1, realistic pallet textures were used and the distance range, as well as the altitude range of the camera were reduced. However, this did not lead to the expected improvement in the network's performance. Especially, in absence of artificial light, the time ratio of successful pallet detection is poor. A different approach was chosen for the generation of the NDDS3 dataset. Distractors, camera wiggle, and additional light sources induced more randomization into the dataset. The results from the experiments prove this approach successful. In all three experiments, NDDS3 achieved a good position accuracy of less than 20cm and a robustness of more than 94%. There are some outliers and some sections where the pose estimation error is large. Applying a low-pass filter has proven to be beneficial in further improving the position accuracy. The filter parameters could be tuned to match the real positions even better. In all three challenging scenes shown in Figure 9 to Figure 11, DOPE trained with NDDS3 performs substantially better than for the other datasets. This, together with the insights gained from the quantitative analysis, underscores the importance of appropriately designing a synthetic data set to cope with challenges such as specific lighting conditions, occlusions, and backgrounds.

Different limitations exist in the datasets, in the experimental evaluation, and the usability of the algorithm. First, the presented datasets are limited in their variability and complexity. Only pallets can be detected. A deviation in the shape and size of a pallet compared to the underlying 3D model can lead to failed detection or incorrect pose estimation. Furthermore, the camera angle and distance to the pallet do not change significantly and the pallet is not occluded by other objects. Therefore, detection and localization at close range and with stacked or loaded pallets is difficult. The experiments were designed accordingly. They were performed in a test hall under laboratory conditions and do not represent a realistic scenario. Considering a real-world application for both dataset generation and experimental design could add value to the results. Another major limitation for practical application of neural networks for pose estimation are the high computational costs, especially when a larger number of objects need to be tracked in real-time. To allow the future application of such computationally intensive algorithms, the trend towards increasing computing power must continue.

## 6 CONCLUSIONS AND OUTLOOK

Various logistics use cases, especially in the field of robotics and automation, require knowledge of the 6D pose of logistics objects. Camera and computer vision technology enable device-free pose estimation, but numerous annotated images are required for training machine learning algorithms. Synthetic training data can be used to avoid the labor-intensive process of manually creating and annotating image datasets. The presented work deals with the RGB camera-based 6D pose estimation of pallets. An analysis of the related work revealed research gaps for the investigation of RGB camera-based 6D pose estimation algorithms and synthetic training datasets for logistics. This work contributed to filling these gaps by providing (1) a photorealistic model of a Euro pallet and a synthetic RGB image dataset including annotations for training 6D pallet pose estimation algorithms, (2) a pipeline to create a neural network trained on synthetic data for 6D pose estimation with an RGB camera, and (3) a qualitative and quantitative evaluation of the position accuracy and robustness of the trained neural network based on different datasets under varying lighting conditions.

Three image datasets were generated by using the *NVIDIA Deep learning Dataset Synthesizer*. The datasets were evaluated, based on three dynamic experiments under different lighting conditions. DOPE trained with the NDDS3 dataset shows superior performance, reaching a mean position error of less than 20cm and an object detection ratio of more than 94% in all three experiments. The qualitative and quantitative results of the different datasets and experiments show how certain lighting conditions, occlusion, and complex backgrounds impose challenges for the object detection that need to be addressed by the training data. Diversifying the images in the dataset by domain randomization and by inducing distractors and camera wiggle is crucial.

For the practical use of deep learning-based 6D pose estimation in logistics and production environments, numerous technical as well as regulatory challenges have to be overcome in terms of data privacy, computational costs, flexibility towards object classes, and robustness. Despite current challenges, the combination of camera and machine learning technology is still the most promising approach when it comes to device-free 6D pose estimation. This work provides a glimpse of what will be possible in logistics and production as deep learning algorithms continue to improve while powerful computers become more affordable.

**Markus Knitt B. Sc.**, Student at the Hamburg University of Technology. During his master's studies in mechatronics, Markus Knitt specialized in robotics and intelligent systems.

**Jakob Schyga M. Sc.**, Research Assistant at the Institute for Technical Logistics, Hamburg University of Technology. Jakob Schyga studied mechanical engineering and production at the Hamburg University of Technology.

**Asan Adamanov M. Sc.**, Research Assistant at the Institute for Technical Logistics, Hamburg University of Technology. Asan Adamanov studied mechanical engineering at the Hamburg University of Technology and Istanbul Technical University.

**Dr. Johannes Hinckeldeyn**, Senior engineer at the Institute for Technical Logistics, Hamburg University of Technology. After completing his doctorate in Great Britain, Johannes Hinckeldeyn worked as Chief Operating Officer for a manufacturer of measurement and laboratory technology for battery research. Johannes Hinckeldeyn studied industrial engineering, production technology, and management in Hamburg and Münster.

**Prof. Dr.-Ing. Jochen Kreutzfeldt**, Professor and Head of the Institute for Technical Logistics, Hamburg University of Technology. After studying mechanical engineering, Jochen Kreutzfeldt held various managerial positions at a company group specializing in automotive safety technology. Jochen Kreutzfeldt then took on a professorship for logistics at the Hamburg University of Applied Sciences and became head of the Institute for Product and Production Management

Address: Institute for Technical Logistics, Hamburg University of Technology, Theodor-Yorck-Strasse 8, 21079 Hamburg, Germany; Phone: +49 40 42878-3557, E-Mail: markus.knitt@tuhh.de